\documentclass{article}

	\PassOptionsToPackage{numbers}{natbib}
     \usepackage[final]{neurips_2020}

\usepackage[utf8]{inputenc} 
\usepackage[T1]{fontenc}    
\usepackage{hyperref}       
\usepackage{url}            
\usepackage{booktabs}       
\usepackage{amsmath, amsfonts}       
\usepackage{nicefrac}       
\usepackage{microtype}      
\usepackage{algorithm, algorithmic}
\usepackage{graphicx}
\usepackage{subfigure}
\usepackage{multirow}
\usepackage{setspace}

\title{Supervised Linear Dimension-Reduction Methods: \\ Review, Extensions, and Comparisons \thanks{The views expressed in the paper are those of the authors and do not represent the views of Wells Fargo.}}

\author{%
  Shaojie~Xu \\
  Wells Fargo \& Company\\
  \texttt{shaojie.xu@wellsfargo.com} \\
  \And
  Joel~Vaughan \\
  Wells Fargo \& Company\\
  \texttt{joel.vaughan@wellsfargo.com} \\
  \And
  Jie~Chen \\
  Wells Fargo \& Company\\
  \texttt{jie.chen@wellsfargo.com} \\
  \And
  Agus~Sudjianto\\
  Wells Fargo \& Company\\
  \texttt{augs.sudjianto@wellsfargo.com} \\ 
  \And
  Vijayan~Nair \\
  Wells Fargo \& Company\\
  \texttt{vijayan.nair@wellsfargo.com} \\
}

\newcommand{\R}{\mathbb{R}}

\renewcommand{\P}[1]{\operatorname{P}\left(#1\right)}

\newcommand{\normal}{\mathcal{N}}

\newcommand{\vct}[1]{\boldsymbol{#1}}
\newcommand{\mtx}[1]{\boldsymbol{#1}}

\newcommand{\<}{\langle}
\renewcommand{\>}{\rangle}

\newcommand{\set}[1]{\mathcal{#1}}

\newcommand{\vb}{\vct{b}}

\newcommand{\vu}{\vct{u}}
\newcommand{\vv}{\vct{v}}

\newcommand{\vx}{\vct{x}}
\newcommand{\vy}{\vct{y}}
\newcommand{\vz}{\vct{z}}
\newcommand{\valpha}{\vct{\alpha}}
\newcommand{\vbeta}{\vct{\beta}}

\newcommand{\vepsilon}{\vct{\epsilon}}

\newcommand{\vtheta}{\vct{\theta}}
\newcommand{\vzero}{\vct{0}}

\newcommand{\mA}{\mtx{A}}

\newcommand{\mC}{\mtx{C}}

\newcommand{\mU}{\mtx{U}}

\newcommand{\mX}{\mtx{X}}

\newcommand{\mZ}{\mtx{Z}}

\newcommand{\mPhi}{\mtx{\Phi}}

\newcommand{\mId}{{\bf I}}

\newcommand{\setV}{\set{V}}

\begin{document}

\maketitle

\setstretch{1.2}

\begin{abstract}
Principal component analysis (PCA) is a well-known linear dimension-reduction method that has been widely used in data analysis and modeling. It is an unsupervised learning technique that identifies a suitable linear subspace for the input variable that contains maximal variation and preserves as much information as possible. PCA has also been used in prediction models where the original, high-dimensional space of predictors is reduced to a smaller, more manageable, set before conducting regression analysis. However, this approach does not incorporate information in the response during the dimension-reduction stage and hence can have poor predictive performance. To address this concern, several supervised linear dimension-reduction techniques have been proposed in the literature. This paper reviews selected techniques, extends some of them, and compares their performance through simulations. Two of these techniques, partial least squares (PLS) and least-squares PCA (LSPCA), consistently outperform the others in this study.
\end{abstract}

\section{Introduction} \label{sec:intro}
Dimension reduction has been widely used in exploratory data analysis (EDA) and as a pre-processing step in high-dimensional prediction problems. In EDA, dimension reduction is used to discover the underlying structure in a dataset: whether the data lie in a smaller subspace so that we can interpret them more easily. In prediction problems, the goal of dimension reduction is to reduce the high-dimensional space of the predictors before fitting a regression model. The rationale is that conventional classification and regression methods often suffer from the “curse of dimensionality”, which states the number of data points required for learning grows exponentially with the dimensionality of the data. Dimension reduction is related to, but different from variable selection, but both share the goal of developing parsimonious and interpretable prediction models. 

Principal component analysis (PCA) is among the earliest dimension-reduction techniques \cite{pearson1901liii, hotelling1933analysis}. It identifies the linear subspace that best approximates the original dataset in terms of retaining the maximum variation. The linear nature of PCA makes the results intrinsically interpretable, as the transformed variables are weighted combinations of the original ones and are orthogonal. 

 PCA has also been used in the context of predictive modeling, where it is called Principal Component Regression (PCR) since it uses PCA to first reduce the dimension and then do regression. However, since the information in the responses is not used during dimension reduction, there is no guarantee that PCR effectively identifies the ``best’’ subspace. In practice, the subspace containing the maximal data variation can differ substantially from the subspace that is most relevant for the prediction problem. 

To address this concern, multiple supervised dimension-reduction techniques have been proposed to take advantage of the information in responses \cite{bair2006prediction, piironen2018iterative, yu2006supervised, barshan2011supervised, geladi1986partial, ritchie2019supervised}. This paper provides a review of these selected approaches and proposes some minor extensions. We restrict our attention to continuous responses. 

The rest of the paper is structured as follows: In Section \ref{sec:preliminaries}, we review PCA and introduce the notation to be used throughout the paper. We divide supervised linear dimension-reduction techniques into two categories. First, Section \ref{sec:wrapper} discusses the class of ``wrapper methods'' that includes methods built around the classical PCA and contain it as a subroutine. Section \ref{sec:intrinsic} discusses the second category named ``intrinsic methods''. These techniques incorporate the label information directly into the objective function of the subspace learning problem. 

We compare the performance of all the algorithms on synthetic and real data sets. The experimental results and analysis are provided in Section \ref{sec:test_sim} and Section \ref{sec:test_real}, using simulated data sets and real data sets respectively. Section \ref{sec:conclusion} concludes the paper with comments on non-linear and semi-supervised dimension reduction techniques.

\section{Preliminaries: PCA} \label{sec:preliminaries}
Consider an unsupervised learning problem where the data are represented in matrix form $\mX \in \R^{N\times P}$. Here, each row of $\mX$ contains one of the $N$ data points, and point $\vx_i \in \R^P$ is a $P-$dimensional vector of variables. Throughout, we assume the data have been column-centered, so that the mean over each column of $\mX$ is zero. The goal in PCA is to find an appropriate linear subspace of $\mX$ and the corresponding orthogonal basis, which can be done by computing the eigenvalue spectral decomposition of the matrix $\mC = \mX^T\mX$. The algorithm is summarized in Algorithm \ref{algo:pca}. In practice, the PCA basis vectors can be calculated directly using singular value decomposition (SVD) on $\mX$.

PCA has several attractive properties. First, the K-dimensional linear subspace spanned by the principle components (PCs), $\vu_1, \vu_2, \ldots, \vu_K$, captures the maximum variability in the data. Second, the transformed variable vectors $\mZ = \mX\mU$ are uncorrelated:
\begin{equation}
\mC_{\mZ} = \mZ^T\mZ = \text{diag}(\lambda_1, \lambda_2, \ldots, \lambda_K) \ ,
\end{equation}
where $\lambda_1, \lambda_2, \ldots, \lambda_K$ are the eigenvalues of $\mC$. Last, the reconstructed data $\hat{\mX} = \mX\mU\mU^T$ are also the best linear approximations of the original data measured by Euclidean norm. That is, $\mU_{\text{PCA}}$ is the maximizer of the following optimization problem:
\begin{equation} \label{eq:pca_opt}
\begin{split}
&\min_{\mU^T\mU=\mId}\ \sum_{i=1}^N ||\vx_i -  \mU\mU^T\vx_i||_2^2 \\
=&\min_{\mU^T\mU=\mId}\ ||\mX - \mX\mU\mU^T||_F^2 \ .
\end{split}
\end{equation}

In the rest of the paper, we focus on the prediction problem where the continuous responses $\vy \in \R^N$ associated with $\mX$ are available. We use $\mU$ to represent the basis. We will add the algorithm name as a subscript when the context is not clear. In addition, we will use superscripts to denote variable selection or iteration. In the presence of a superscript, the inner product $\vx^T\vy$ will be written as $\<\vx, \vy\>$.

\begin{algorithm}[H]
\caption{PCA} \label{algo:pca}
\begin{algorithmic}
\STATE Given centered training data $\mX \in \R^{N\times P}$ and a desired lower dimension $K$.
\STATE Calculate the covariance matrix $\mC = \mX^T\mX$.
\STATE Calculate the first $K$ largest eigenvalues of $\mC$ and their corresponding eigenvectors $\vu_1, \vu_2, \ldots, \vu_K$.
\STATE Build the PCA basis $\mU \in \R^{P\times K}$, $\mU = [\vu_1, \vu_2, \ldots, \vu_K]$.
\STATE \textbf{To transform a new sample} $\vx^*$:
	\begin{ALC@g}
	\STATE Return $\vz^* = \mU^T\vx^*$
    \end{ALC@g}
\end{algorithmic}
\end{algorithm}
\vspace{-3mm}

\section{Wrapper methods} \label{sec:wrapper}
We first consider the wrapper methods that contain PCA as a subroutine and build around it to incorporate the label information. Since PCA learns the subspace in an unsupervised manner, the resulting subspace may not be the ``best’’ in developing a regression model. Supervised linear dimension-reduction methods address this concern by incorporating the response information into the learning process. One way to utilize the information is to add pre-processing/post-processing steps before/after performing PCA.

\subsection{Bair method} 
Bair et al. \cite{bair2006prediction} proposed a method that adds a variable selection step before running PCA. We call this method Bair method in our paper. It selects a subset $M$ of the original $P$ variables that are the most related to the responses and does PCA on this reduced variable matrix. Define a score function $s(\vx^j, \vy)$ that measures how closely the $j$th variable is associated with the responses. One choice for the score function is the absolute value of the covariance:
\begin{equation} \label{eq:cov_score}
s(\vx^j, \vy) = | \< \vx^j , \vy \> | \ .
\end{equation}
However, since the variables may have different variances, it may be better to consider the Pearson correlation coefficient:
\begin{equation} \label{eq:person_corr}
s(\vx^j, \vy) = \frac{| \< \vx^j , \vy \> |}{||\vx^j||_2 ||\vy||_2} \ .
\end{equation} 

We can rank the ``importance’’ of the variables by their scores using either one of the two score functions. We then select the top $M$ variables to form a new data matrix $\tilde{\mX}$ using the corresponding columns from $\mX$. Finally, PCA is performed on the new variable matrix $\tilde{\mX}$. 

\begin{algorithm}[H]
\caption{Bair Method} \label{algo:bair}
\begin{algorithmic}
\STATE Given centered training data $\mX \in \R^{N\times P}$, their corresponding centered label $\vy \in \R^N$, and a desired lower dimension $K$.
\STATE Given a variable score function $s(\vx^j, \vy)$, i.g. (\ref{eq:cov_score}, \ref{eq:person_corr}).
\STATE Calculate the score of each variable: $s_j = s(\vx^j, \vy) $ and rank them in a decreasing order.
\FOR{$M = K, K+1, \ldots, P$}
\STATE Choose the first $M$ variables which form a set $\setV$.
\STATE Construct the new data matrix $\tilde{\mX}$ by selecting the corresponding columns $\vx^j, j\in\setV$, from $\mX$.
\STATE Perform PCA as in Algorithm \ref{algo:pca} on $\tilde{\mX}$. Get the basis $\mU$ and the transform the data $\tilde{\mZ} = \tilde{\mX}\mU$.
\STATE Build the subsequent prediction model using $\tilde{\mZ}$ and $\vy$.
\STATE Record the prediction performance.
\ENDFOR
\STATE Choose the $M$ with the best performance and get the corresponding $\setV$ and $\mU$.
\STATE \textbf{To transform a new sample} $\vx^*$:
	\begin{ALC@g}
	\STATE Form a new sample $\tilde{\vx}^*$ by selecting vairalbes $j \in \setV$.
	\STATE Return $\vz^* = \mU\tilde{\vx}^*$.
    \end{ALC@g}
\end{algorithmic}
\end{algorithm}
\vspace{-3mm}

Bair method contains one hyperparameter $M$, which can be selected based on the performance of the subsequent prediction model measured on the training or validation set. The method is summarized in Algorithm \ref{algo:bair}. In this version, we directly use the performance on the training set to select $M$.

By eliminating the variables that contain little information about the responses, Bair method learns a more ``informative'' linear subspace. Variables with large variation tend to dominate the classic PCA, but they may not be relevant for prediction. Bair method addresses this problem and effectively controls the impact of these variables.

\subsection{Piironen-Vehtari (PV) method} 
An iterative version of Bair method was proposed in \cite{piironen2018iterative} by Piironen and Vehtari. Although it was named ISPCA in the original paper, we refer to it as PV method. It runs Bair algorithm iteratively and gets one PC each time. Under the $k$th iteration, a subset of $M^k$ variables that are most related to the responses are chosen to form a new dataset. Denote the new dataset as $\tilde{\mX}^k$ which contains the selected columns from the full dataset matrix $\mX^k$. Only the first PC is calculated from $\tilde{\mX}^k$. As the transformed variables within each iteration are now one-dimensional, $M^k$ can be chosen so that they are most correlated with the response. 

Denote $\setV^k$ as the set of the selected variables. PV method then subtracts the variation explained by the chosen PC, $\vu^k$, from the data by regressing each feature (both selected and unselected) on the one-dimensional transformed data:
\begin{align}
&\vz^k = \tilde{\mX}^k \vu^k \label{eq:piironen_update_transformed_data}\ , \quad \vb^k = \frac{(\mX^k)^T \vz^k}{||\vz^k||_2^2} \\
&\mX^{k+1} = \mX^{k}  - \vz^k (\vb^k)^T \label{eq:piironen_var_update} \ .
\end{align}

\begin{algorithm}[H]
\caption{Piironen-Vehtari (PV) Method} \label{algo:piironen}
\begin{algorithmic}
\STATE Given centered training data $\mX \in \R^{N\times P}$, their corresponding centered label $\vy \in \R^N$, and a desired lower dimension $K$.
\STATE Given a variable score function $s(\vx^j, \vy)$, i.g. (\ref{eq:cov_score}, \ref{eq:person_corr}).
\STATE Set $\mX^1 = \mX$
\FOR{$k = 1, 2, \ldots, K$}
\STATE Calculate the score of each variable: $s_j = s(\vx^{j,k}, \vy) $ and rank them in a decreasing order.
\FOR{$M^k = 1, 2, \ldots, P$}
\STATE Choose the first $M^k$ variables which form a set $\setV^k$.
\STATE Construct the new data matrix $\tilde{\mX}^k$ by selecting the corresponding columns $\vx^{j, k}, j\in\setV^k$, from $\mX^k$.
\STATE Perform PCA as in Algorithm \ref{algo:pca} on $\tilde{\mX}^k$ to get the first PC, $\vu^k$, and the transformed data, $\vz^k$.
\STATE Calculate the score $s(\vz^k, \vy)$ using (\ref{eq:person_corr}).
\ENDFOR
\STATE Choose the $M^k$ with the highest score and record and corresponding $\setV^k$, $\vu^k$.
\STATE Update the variables and the responses using (\ref{eq:piironen_update_transformed_data}-\ref{eq:piironen_var_update}) and record $\vb^k$.
\ENDFOR
\STATE \textbf{To transform a new sample} $\vx^*$:
	\begin{ALC@g}
	\STATE Set $\vx^1 = \vx^*$
	\FOR{$k = 1, 2, \ldots, K$}	
	\STATE Construct a new sample $\tilde{\vx}^k$ by selecting variables $j \in \setV^k$.
	\STATE Calculate $z^k = \<\vu^k, \tilde{\vx}^k\>$.
	\STATE Update the variables: $\vx^{k+1} = \vx^{k} - z^k \vb^k$.
	\ENDFOR
	\STATE Return $\vz^* = [z^1, z^2, \ldots, z^k]^T$.
    \end{ALC@g}
\end{algorithmic}
\end{algorithm}
\vspace{-3mm}

PV method is summarized in Algorithm \ref{algo:piironen}. Unlike Bair method, it uses an iterative scheme to select a diverse set of variables in each iteration, all of which will contribute to the transformed data. The iterative scheme breaks one of the major properties of the classic PCA: the basis vectors $\vu^k$ obtained by this method are no longer orthogonal to each other. Nevertheless, the features in the transformed space are still uncorrelated with each other: $\<\vz^i,\vz^j\> = 0\ \forall i \neq j$ \cite{piironen2018iterative}.

\subsection{PC post-selection (PCPS) method}
The previous two methods use variable selection as a pre-processing step before applying the classic PCA. We also consider a post-processing step after PCA to incorporate the response information. We refer to it as the PC post-selection method (PCPS). We first conduct PCA with all $P$ variables to obtain the full set of PCs, and then calculates the transformed data along each PC: $\vz_k = \mX\vu_k,\ k=1,2,\ldots,P$. One of the two score functions discussed earlier (\ref{eq:cov_score}, \ref{eq:person_corr}) is then used to measure the relationship between the transformed dimensions and the responses: $s_k = s(\vz_k, \vy)$. The first K PCs corresponding to the highest scores are chosen to form the basis of the learned linear subspace. This technique is summarized in Algorithm \ref{algo:pcs}.

\begin{algorithm}[H]
\caption{PC Post-Selection (PCPS) Method} \label{algo:pcs}
\begin{algorithmic}
\STATE Given centered training data $\mX \in \R^{N\times P}$, their corresponding centered label $\vy \in \R^N$, and a desired lower dimension $K$.
\STATE Given a PC score function $s(\vz_k, \vy)$, i.g. (\ref{eq:cov_score}, \ref{eq:person_corr}).
\STATE Perform PCA as in Algorithm \ref{algo:pca} on the new data matrix, and get the PCs, $\vu_1, \vu_2, \ldots, \vu_P$.
\STATE Calculate the score of each PC: $s_k = s(\mX\vu_k, \vy)$
\STATE Build the basis $\mU$ by selecting the first $K$ PCs with the highest scores. 
\STATE \textbf{To transform a new sample} $\vx^*$:
	\begin{ALC@g}
	\STATE Return $\vz^* = \mU^T\vx^*$
    \end{ALC@g}
\end{algorithmic}
\end{algorithm}
\vspace{-3mm}

\section{Intrinsic methods} \label{sec:intrinsic}
As noted in (\ref{eq:pca_opt}), PCA can be derived from an optimization problem. One can expand the formulation in (\ref{eq:pca_opt}) to incorporate the label information into the objective function, which is the underlying idea of several supervised linear dimension reduction algorithms. Since the objective function contains both supervised and unsupervised components, we call them ``intrinsic methods''.

\subsection{Partial Least Squares (PLS)}
Partial least squares (PLS) has a long history. It was originally proposed in \cite{wold1975soft} and there have been many variations developed since then \cite{geladi1986partial}. Although PLS has been used mainly with multi-variate responses, our review focuses on its application to one-dimensional responses. 

PLS constructs the basis of a linear subspace iteratively. Within each iteration $k$, a basis vector $\vu^k$ is calculated by maximizing the covariance between the responses and the transformed input variables:
\begin{equation}  \label{eq:PLS_iter_obj}
\max_{||\vu^k||=1}\ \left((\mX^k\vu^k)^T\vy^k\right)^2 \ .
\end{equation}
This optimization problem can be solved directly using eigenvalue decomposition, while the iterative solver in the original PLS can be viewed as the power-iteration method for computing the eigenvectors. After each iteration, both the variable matrix $\mX$ and the responses are updated by subtracting the parts explained along the direction of $\vu^k$:
\begin{align} 
& \vz^{k} = \mX^{k}\vu^k\ , \quad \mX^{k+1} = \mX^{k} -\vz^k(\vu^k)^T \label{eq:PLS_iter_update_x}  \\ 
& \vy^{k+1} = \vy^k - \frac{\<\vy^k, \vz^k\>}{||\vz^k||_2^2}\vz^k \ . \label{eq:PLS_iter_update_y}
\end{align}
PLS is described in Algorithm \ref{algo:pls_ori}.

We also consider an extended version of the PLS algorithm by combining the supervised objective in (\ref{eq:PLS_iter_obj}) and the unsupervised objective in PCA (\ref{eq:pca_opt}). The new objective function in each iteration is now:
\begin{equation}  \label{eq:pls_obj_eq_ext}
\max_{||\vu^k||=1}\ \left((\mX^k\vu^k)^T\vy^k\right)^2 - \gamma ||\mX^k - \mX^k\vu^k(\vu^k)^T||_F^2 \ ,
\end{equation}
and an equivalent form is:
\begin{equation}  \label{eq:pls_obj_ext}
\max_{||\vu^k||=1}\ (\vu^k)^T(\mX^k)^T \left(\vy^k(\vy^k)^T+\gamma\mId\right) \mX^k\vu^k .
\end{equation}
The hyperparameter $\gamma$ is used to balance between the supervised and the unsupervised objectives. When $\gamma=0$, the new model reduces to the original PLS method. When $\gamma \to \infty$, the new model becomes the classic PCA. Similar to PCA, (\ref{eq:pls_obj_ext}) can be solved by eigenvalue decomposition. Our extended version of PLS is summarized in Algorithm \ref{algo:pls}.

 \begin{algorithm}[H]
\caption{PLS} \label{algo:pls_ori}
\begin{algorithmic}
\STATE Given centered training data $\mX \in \R^{N\times P}$, their corresponding centered label $\vy \in \R^N$, and a desired lower dimension $K$.
\STATE Set $\mX^1 = \mX$, $\vy^1 = \vy$
\FOR{$k = 1, 2, \ldots, K$}
\STATE Solve (\ref{eq:PLS_iter_obj}) by calculating the largest eigenvalue of $\mC=(\mX^k)^T \vy^k(\vy^k)^T \mX^k$ and setting its corresponding eigenvector as $\vu^k$.
\STATE Update the data matrix and the responses using (\ref{eq:PLS_iter_update_x}-\ref{eq:PLS_iter_update_y}).
\ENDFOR
\STATE Build the basis $\mU \in \R^{P\times K}$, $\mU = [\vu^1, \vu^2, \ldots, \vu^K]$.
\STATE \textbf{To transform a new sample} $\vx^*$:
	\begin{ALC@g}
	\STATE Return $\vz^* = \mU^T\vx^*$
	\end{ALC@g}
\end{algorithmic}
\end{algorithm}
\vspace{-3mm}

 \begin{algorithm}[H]
\caption{Extended PLS} \label{algo:pls}
\begin{algorithmic}
\STATE Given centered training data $\mX \in \R^{N\times P}$, their corresponding centered label $\vy \in \R^N$, and a desired lower dimension $K$.
\STATE Select a value for the hyperparameter $\gamma$.
\STATE Set $\mX^1 = \mX$, $\vy^1 = \vy$
\FOR{$k = 1, 2, \ldots, K$}
\STATE Solve (\ref{eq:pls_obj_eq_ext}) by calculating the largest eigenvalue of $\mC=(\mX^k)^T \left(\vy^k(\vy^k)^T+\gamma\mId\right) \mX^k$ and setting its corresponding eigenvector as $\vu^k$.
\STATE Update the data matrix and the responses using (\ref{eq:PLS_iter_update_x}-\ref{eq:PLS_iter_update_y}).
\ENDFOR
\STATE Build the basis $\mU \in \R^{P\times K}$, $\mU = [\vu^1, \vu^2, \ldots, \vu^K]$.
\STATE \textbf{To transform a new sample} $\vx^*$:
	\begin{ALC@g}
	\STATE Return $\vz^* = \mU^T\vx^*$
	\end{ALC@g}
\end{algorithmic}
\end{algorithm}
\vspace{-3mm}

\subsection{Barshan method}
Barshan et. al. formulated their method \cite{barshan2011supervised} in a general manner using reproducing kernel Hilbert space. We call this method Barshan method in our paper. The ``kernel trick'' allows this method to achieve non-linear dimension reduction. This level of generality is beyond the scope of this review paper. We focus on a specific version of Barshan method with a linear target variable (response) kernel. The objective function of this version has the following form:
\begin{equation}  \label{eq:barshan_obj}
\max_{\mU^T\mU = \mId}\ \text{trace} \left( \mU^T \mX^T \vy\vy^T \mX\mU \right) \ .
\end{equation}
It can also be represented in an equivalent form:
\begin{equation}  \label{eq:barshan_obj_eq}
\max_{\mU^T\mU=\mId}\ ||\left(\mX\mU\right)^T\vy||_2^2 \ .
\end{equation}
Equation (\ref{eq:barshan_obj_eq}) can be viewed as maximizing the sum of squares of the covariance between the response and each dimension of the transformed variable $\mZ = \mX\mU$. The method is described in Algorithm \ref{algo:barshan_ori}.

Notice that the formulation in (\ref{eq:barshan_obj_eq}) becomes (\ref{eq:PLS_iter_obj}) when $\mU$ contains only one column. Therefore, PLS can be viewed as running Barshan method with a linear target variable kernel within each iteration \cite{barshan2011supervised}. The basis learned by Algorithm \ref{algo:barshan_ori} and \ref{algo:pls_ori} share the same first basis vector. However, the rest of the vectors are different and they span different subspaces. Both bases are orthogonal.

Similar to how we extended PLS to include an unsupervised objective, we can also modify Barshan method to include an unsupervised component. The new objective function is now:
\begin{equation}  \label{eq:barshan_obj_eq_ext}
\max_{\mU^T\mU=\mId}\ ||\left(\mX\mU\right)^T\vy||_2^2 - \gamma ||\mX - \mX\mU\mU^T||_F^2 \ ,
\end{equation}
with its equivalent form:
\begin{equation}  \label{eq:barshan_obj_ext}
\max_{\mU^T\mU = \mId}\ \text{trace} \left( \mU^T\mX^T (\vy\vy^T+\gamma\mId) \mX\mU \right) \ .
\end{equation}
Our extended version of Barshan method is summarized in Algorithm \ref{algo:barshan}.

\begin{algorithm}[H]
\caption{Barshan Method with Linear Target Variable Kernel} \label{algo:barshan_ori}
\begin{algorithmic}
\STATE Given centered training data $\mX \in \R^{N\times P}$, their corresponding centered label $\vy \in \R^N$, and a desired lower dimension $K$.
\STATE Solve (\ref{eq:barshan_obj}) by calculating the first $K$ largest eigenvalues of $\mC=\mX^T \vy\vy^T \mX$ and their corresponding eigenvectors $\vu_1, \vu_2, \ldots, \vu_K$.
\STATE Build the PCA basis $\mU \in \R^{P\times K}$, $\mU = [\vu_1, \vu_2, \ldots, \vu_K]$.
\STATE \textbf{To transform a new sample} $\vx^*$:
	\begin{ALC@g}
	\STATE Return $\vz^* = \mU^T\vx^*$
	\end{ALC@g}
\end{algorithmic}
\end{algorithm}
\vspace{-3mm}

\begin{algorithm}[H]
\caption{Extended Barshan Method with Linear Target Variable Kernel} \label{algo:barshan}
\begin{algorithmic}
\STATE Given centered training data $\mX \in \R^{N\times P}$, their corresponding centered label $\vy \in \R^N$, and a desired lower dimension $K$.
\STATE Select a value for the hyperparameter $\gamma$.
\STATE Solve (\ref{eq:barshan_obj_ext}) by calculating the first $K$ largest eigenvalues of $\mC=\mX^T (\vy\vy^T+\gamma\mId) \mX$ and their corresponding eigenvectors $\vu_1, \vu_2, \ldots, \vu_K$.
\STATE Build the PCA basis $\mU \in \R^{P\times K}$, $\mU = [\vu_1, \vu_2, \ldots, \vu_K]$.
\STATE \textbf{To transform a new sample} $\vx^*$:
	\begin{ALC@g}
	\STATE Return $\vz^* = \mU^T\vx^*$
	\end{ALC@g}
\end{algorithmic}
\end{algorithm}
\vspace{-3mm}

\subsection{Least-squares PCA (LSPCA)}
In Barshan method, the supervised objective measures the sum of the covariance between the responses and the transformed variables along each direction. An alternative is to measure the least square error of regressing the responses onto these variables. This approach, named least-squares PCA (LSPCA) was proposed by \cite{ritchie2019supervised}. It learns a linear subspace by solving the following optimization problem:  
\begin{equation} \label{eq:lspca_obj}
\min_{\mU^T\mU = \mId,\vbeta}\ ||\vy - \mX \mU \vbeta||_2^2 + \gamma ||\mX - \mX \mU \mU^T||_F^2 \ .
\end{equation}

Unlike Barshan method, this optimization problem does not have a known closed-form solution. One method to solve it, based on matrix manifold optimization, is proposed in \cite{ritchie2019supervised}. Other optimization methods with orthogonal constraints have been discussed in \cite{absil2009optimization, wen2013feasible}. We refer readers to Algorithm 1 in the original paper \cite{ritchie2019supervised} for the details.

We have already noted the similarities between the objective functions of Barshan method (\ref{eq:barshan_obj_eq_ext}) and LSPCA (\ref{eq:lspca_obj}). It can be shown that the two are indeed equivalent if $\mX^T\mX = \mId$, that is, when the data have been standardized and decorellated. In this scenario, the optimal $\vbeta$ for any given $\mU$ has the form:
\begin{equation}
\vbeta^* = (\mU^T\mX^T\mX\mU)^{-1}\mU^T\mX^T\vy = \mU^T\mX^T\vy \ .
\end{equation}
By plugging $\vbeta^*$ back into the first term in (\ref{eq:lspca_obj}), we have:
\begin{equation}
\begin{split}
&\min_{\mU^T\mU=\mId}\  ||\vy - \mX \mU \vbeta^*||_2^2 \\
=& \min_{\mU^T\mU=\mId}\  ||\vy - \mX \mU \mU^T\mX^T\vy||_2^2 \\
=& ||\vy||_2^2 + \min_{\mU^T\mU=\mId}\  - \vy^T\mX \mU \mU^T\mX^T\vy \\
=& ||\vy||_2^2 - \max_{\mU^T\mU=\mId}\  |||\left(\mX\mU\right)^T\vy||_2^2 \ .
\end{split}
\end{equation}
Therefore, Barshan method in Algorithm \ref{algo:barshan} and LSPCA learn the same subspace when $\mX^T\mX = \mId$.

\subsection{Supervised probabilistic PCA (SPPCA) method} 
The SPPCA method \cite{yu2006supervised} is a model-based approach. It assumes that both the high-dimensional observed input variables and their responses are controlled by some low dimensional latent variables and are conditionally independent given these latent variables. Specifically,
\begin{align}
\vx &= \mU\vz + \vepsilon_x \ , \\
y &= \vv^T\vz + \epsilon_y \ ,
\end{align}
where $\vz\in\R^K$, $\vx\in\R^P$, $y\in\R$, $\vepsilon_x\in\R^P$, $\epsilon_y\in\R$ are random variables following Gaussian distributions: $\vz \sim \normal (\vzero, \mId)$, $\vepsilon_x \sim \normal (\vzero, \sigma_x^2\mId)$, $\epsilon_y \sim \normal (\vzero, \sigma_y^2\mId)$. This probabilistic model contains four parameters: $\vtheta$ =  $\{\mU, \vv, \sigma_x, \sigma_y\}$.

Given $N$ observation pairs $(\vx_i, y_i)$, we can obtain the maximum likelihood estimates (MLEs) of the model parameters by solving:
\begin{equation}
\begin{split}
&\max_{\vtheta}\ \prod_{i=1}^N \P{\vx_i, y_i; \vtheta} \\
=& \max_{\vtheta}\ \prod_{i=1}^N \int \P{\vx_i|\vz; \vtheta}\P{y_i|\vz; \vtheta}\P{\vz; \vtheta} d\vz
\end{split}
\end{equation}
We refer readers to Algorithm 1 in the original paper \cite{yu2006supervised} for an expectation-maximization (EM) algorithm to obtain the MLE. \footnote{The term $\mA^{-1}$ in equation (6) in the original paper should have been multiplied by the number of samples.}

As the response is assumed to have a linear relationship with the latent variable, SPPCA returns a dimension reduction scheme and a prediction model simultaneously. However, the basis $\mU$ obtained by this method is no longer orthogonal. Given a new sample $\vx^*$, the predicted value of the corresponding latent variable is
\begin{equation}
\vz^* = (\mU^T\mU+\sigma_x^2\mId)^{-1}\mU^T\vx^* \ ,
\end{equation} 
and its predicted label can be calculated as $y^* = \vv^T\vz^*$.  

SPPCA maximizes the likelihood of the data-label pair jointly and its performance is often sensitive to the dimension of the data \cite{ritchie2019supervised}. High dimensional data tend to overpower the information provided by the response, leading to results that are very similar to the classic PCA.

\section{Comparison of the methods using simulated data sets} \label{sec:test_sim}
\begin{figure}[htb]
\center
\includegraphics[width=.75\columnwidth]{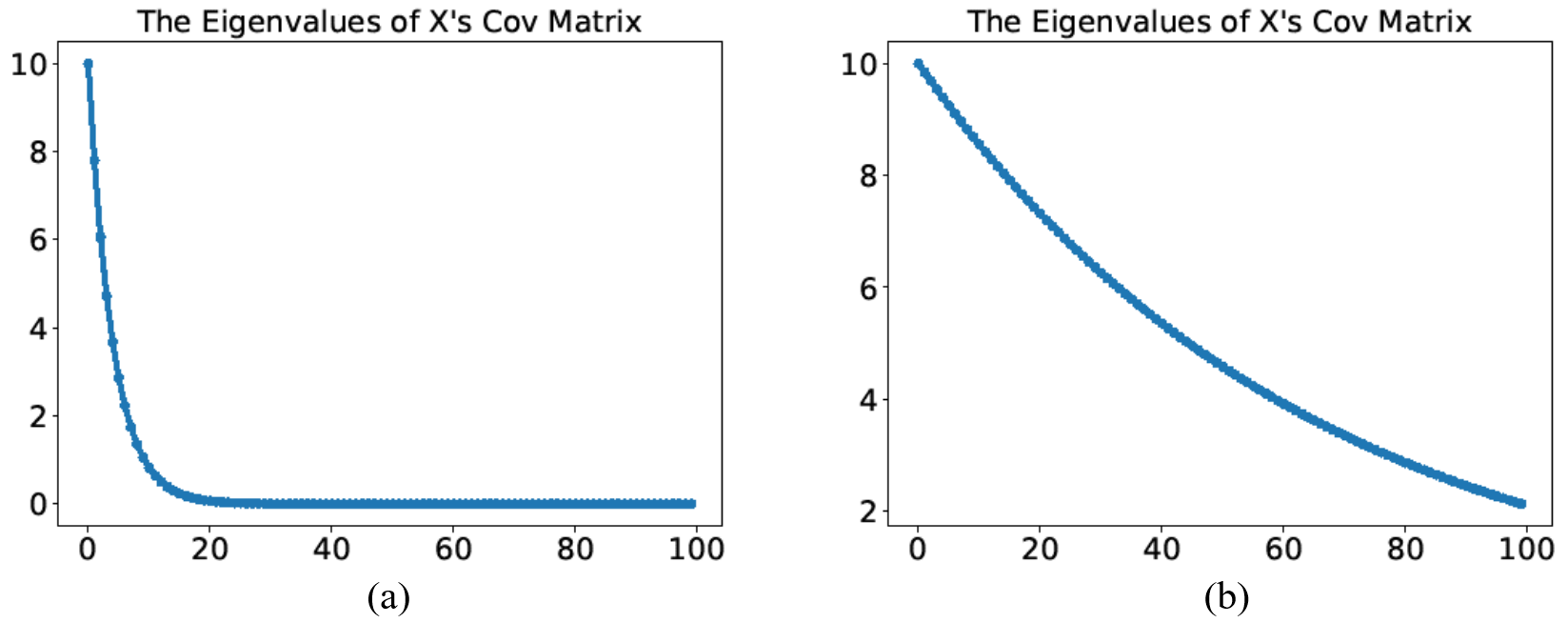}
\vspace{-3mm}
\caption{The eigenvalues of X's covariance matrix, (a) the case of fast decay and (b) the case of slow decay.}
\vspace{-3mm}
\label{fig:eigvals_cov_x}
\end{figure}

We compared the performance of the reviewed algorithms using simulations. For Barshan and PLS, the experiments were based on the extended versions rather than the original ones. The simulated data $\vx \in R^{100}$ were generated by i.i.d. sampling from multivariate Gaussian. We examined two different sample sizes: $N_{train} \in \{150, 1500\}$. The first case with $N = 150$ and $P = 100$ revealed potential problems when there was “curse-of-dimensionality”. We simulated $10,000$ samples for testing. We designed different scenarios for the distribution of the the eigenvalues of $\mC = \mX^T\mX$. As shown in Figure \ref{fig:eigvals_cov_x}, the decay of the eigenvalues in the first case was exponential in nature and was much faster than that in the second case which was close to linear. 

The relationship between the input variables and their responses satisfied a linear model:
\begin{equation}
\vy = \mX\vbeta + \vepsilon = \mX\mPhi\valpha + \vepsilon \ ,
\end{equation}
where the coefficient vector $\vbeta \in \R^{100}$ was confined in a 10-dimensional linear subspace spanned by the columns of $\mPhi\in\R^{100\times10}$ with its corresponding lower-dimensional representation as $\valpha \in \R^{10}$. We fixed $\valpha$ as a vector of all $1$s. The term $\vepsilon$ was an additive Gaussian noise with standard deviation of $0.5$ when the eigenvalues of $\mC$ decayed fast and was $2.5 $ when the decay was slow. 

We designed three cases to cover different alignments between the subspace containing the maximal data variation and the subspace where the coefficient $\vbeta$ lived. In the ``well-aligned'' case, we constructed $\mPhi$ such that its columns were the eigenvectors of $\mC$ corresponding to the top ten largest eigenvalues. In the ``misaligned'' case, the columns of $\mPhi$ were the 11th to the 20th eigenvectors. In the ``partially-aligned'' case, we included the 11th, 13th, 15th, 17th, and 19th eigenvectors in $\mPhi$, and also added another 5 randomly chosen vectors that were orthogonal to the 5 included eigenvectors. The randomly chosen vectors tended to capture some of the space spanned by top vectors.

Since the algorithms did not know the true dimension of the linear embedding, we chose to learn a $15$-dimensional subspace. The $\gamma$ hyperparameter in Barshan method, PLS, and LSPCA was tuned on a validation set. After the dimension-reduction step, we built a regression model using the transformed input variables and responses. We also included a baseline algorithm which ran ordinary least square (OLS) directly on the original high-dimensional data. For each test case, we ran 100 trials. In each trial, the eigenvectors of the covariance matrix $\mC$ were randomly constructed, and the training and testing data were randomly drawn from the distribution. The mean squared errors (MSEs) of each algorithm were calculated on the training and testing samples in each trial. We used the average MSEs over the 100 trials to compare the performance of different models.

\subsection{Results of the case with fast decaying eigenvalues in $\mC$}
We first discuss the results of the case with fast decaying eigenvalues as shown in Figure \ref{fig:eigvals_cov_x}.a. The average training and testing MSEs are summarized in Table \ref{tab:sim_res_fastdecay}. The top performing algorithms under each testing setting are highlighted. When the number of training samples is limited, OLS severely overfits the data as it does not exploit the low-dimensional embedding. However, the classic PCA performs well when there is a good alignment between the subspace containing the maximal data variation and the subspace where the coefficient $\vbeta$ lives. The tuned $\gamma$s in Barshan method, PLS, and LSPCA have large values, so the algorithms behave the same as the classic PCA. PV and PCPS methods show some sign of overfitting. 

When the subspaces are not well-aligned, the classic PCA uses a ``unsatisfactory'' subspace for the prediction task, so the performance is much worse. As discussed earlier, the results of SPPCA are very similar to the PCA, since the information contained in the high-dimensional data themselves has overpowered the information provided by the responses. Bair method, Barshan method, and PCPS all find better subspace by exploiting the information in the responses. The best performance is achieved by PV, PLS, and LSPCA. Combining with the results from the first simulation test, we see that the flexible tuning scheme allows LSPCA and PLS to consistently achieve the best performance.
 
\subsection{Results of the case with slowly decaying eigenvalues in $\mC$}
We repeated the simulation tests with new data where the covariance matrix $\mC$ has slowly decaying eigenvalues as shown in  Figure \ref{fig:eigvals_cov_x}.b. The average training and testing MSEs are summarized in Table \ref{tab:sim_res_slowdecay}. Many of the findings are similar to those in the previous case, with the main difference being for PCA. When the eigenvalues of $\mC$ decay slowly, PCA is unable to achieve good performance even when the subspace spanned by the top PCs is well-aligned with the subspace in which $\vbeta$ lives, especially when the number of training samples is small. A closer investigation shows that the limited number of training samples causes relatively large error in the estimation of the eigenvectors. Additionally, the lack of significant differences among the top eigenvalues manifests the impact of these estimation errors by causing PCA to choose the wrong subspace. Supervised dimension-reduction algorithms use the information in the responses to alleviate this problem. LSPCA and PLS  continue to achieve top performance.

\begin{table}[ht]
\centering
\small
\renewcommand{\arraystretch}{1.5}
\begin{tabular}{  c | c | c | c | c | c | c  }
    \hline\hline
    & \multicolumn{2}{c|}{Well-Aligned Subspaces} & \multicolumn{2}{c|}{Misaligned Subspaces} & \multicolumn{2}{c}{Partially-Aligned Subspaces} \\ \hline
    $N_{train}=$ & 150 & 1500 & 150 & 1500 & 150 & 1500 \\ \hline 
    \hline
    OLS & 0.082 / 0.763 & 0.233 / 0.267 & 0.081 / 0.781 & 0.234 / 0.267 & 0.081 / 0.771 & 0.234 / 0.268 \\ \hline 
    PCA & 0.226 / \textbf{0.285} & 0.248 / 0.253 & 0.875 / 1.154 & 0.990 / 1.019 & 0.538 / 0.700 & 0.583 / 0.595 \\ \hline 
    Bair & 0.222 / 0.287 & 0.248 / 0.253 & 0.548 / 0.741 & 0.552 / 0.567 & 0.404 / 0.544 & 0.464 / 0.476 \\ \hline 
    PV & 0.206 / 0.316 & 0.253 / 0.262 & 0.216 / \textbf{0.330} & 0.261 / 0.270 & 0.209 / \textbf{0.320} & 0.258 / \textbf{0.267} \\ \hline 
    PCPS & 0.204 / 0.370 & 0.246 / 0.255 & 0.247 / 0.356 & 0.253 / \textbf{0.259} & 0.263 / 0.376 & 0.286 / 0.292 \\ \hline 
    Barshan & 0.224 / \textbf{0.285} & 0.248 / \textbf{0.252} & 0.508 / 0.689 & 0.544 / 0.560 & 0.423 / 0.560 & 0.465 / 0.475 \\ \hline 
    PLS & 0.216 / \textbf{0.284} & 0.247 / \textbf{0.252} & 0.237 / \textbf{0.338} & 0.257 / \textbf{0.264} & 0.231 / \textbf{0.323} & 0.262 / \textbf{0.268} \\ \hline 
    LSPCA & 0.216 / 0.285 & 0.247 / \textbf{0.252} & 0.192 / \textbf{0.315} & 0.246 / \textbf{0.254} & 0.193 / \textbf{0.311} & 0.248 / \textbf{0.257} \\ \hline 
    SPPCA & 0.225 / 0.285 & 0.248 / 0.253 & 0.810 / 1.074 & 0.911 / 0.939 & 0.510 / 0.666 & 0.554 / 0.566 \\ \hline 
    \hline
\end{tabular}
\vspace{3mm}
\caption{The average training/testing MSEs from 100 simulation trials under different settings. The eigenvalues of $\mC$ decay fast. The top three testing MSEs under each setting are in bold.}
\label{tab:sim_res_fastdecay}
\vspace{-5mm}
\end{table}

\begin{table}[ht]
\centering
\small
\renewcommand{\arraystretch}{1.5}
\begin{tabular}{  c | c | c | c | c | c | c  }
    \hline\hline
    & \multicolumn{2}{c|}{Well-Aligned Subspaces} & \multicolumn{2}{c|}{Misaligned Subspaces} & \multicolumn{2}{c}{Partially-Aligned Subspaces} \\ \hline
    $N_{train}=$ & 150 & 1500 & 150 & 1500 & 150 & 1500 \\ \hline 
    \hline
    OLS & 2.05 / 19.79 & 5.81 / \textbf{6.70} & 2.05 / 19.23 & 5.83 / \textbf{6.69} & 2.04 / \textbf{19.28} & 5.84 / \textbf{6.68} \\ \hline 
    PCA & 33.60 / 52.77 & 22.41 / 24.41 & 38.72 / 57.36 & 45.19 / 49.57 & 37.04 / 52.11 & 43.04 / 46.46 \\ \hline 
    Bair & 17.40 / 32.83 & 11.59 / 12.50 & 18.93 / 34.97 & 14.38 / 15.73 & 18.53 / 34.59 & 19.54 / 21.40 \\ \hline 
    PV & 4.67 / \textbf{17.14} & 6.72 / 7.55 & 4.88 / \textbf{17.88} & 7.68 / 8.66 & 5.32 / 19.64 & 9.03 / 10.15 \\ \hline 
    PCPS & 18.51 / 32.58 & 12.90 / 13.85 & 20.11 / 34.74 & 16.75 / 17.95 & 19.83 / 34.18 & 23.71 / 25.35 \\ \hline 
    Barshan & 11.17 / 24.44 & 7.27 / 7.96 & 11.71 / 25.59 & 7.45 / 8.25 & 12.10 / 26.28 & 10.44 / 11.55 \\ \hline 
    PLS & 2.77 / \textbf{13.02} & 5.81 / \textbf{6.70} & 2.86 / \textbf{13.18} & 5.83 / \textbf{6.69} & 2.90 / \textbf{14.65} & 5.84 / \textbf{6.68} \\ \hline 
    LSPCA & 3.03 / \textbf{13.16} & 5.86 / \textbf{6.66} & 3.02 / \textbf{13.19} & 5.83 / \textbf{6.69} & 2.98 / \textbf{14.49} & 5.84 / \textbf{6.68} \\ \hline 
    SPPCA & 12.09 / 27.30 & 13.73 / 15.30 & 12.41 / 27.74 & 14.26 / 15.94 & 12.31 / 27.22 & 15.05 / 16.75 \\ \hline 
    \hline
\end{tabular}
\vspace{3mm}
\caption{The average training/testing MSEs from 100 simulation trials under different settings. The eigenvalues of $\mC$ decay slowly. The top three testing MSEs under each setting are in bold.}
\label{tab:sim_res_slowdecay}
\vspace{-5mm}
\end{table}

\subsection{The effect of $\gamma$ on LSPCA and on the extensions of Barshan and PLS}
\begin{figure}[htb]
\center
\includegraphics[width=\columnwidth]{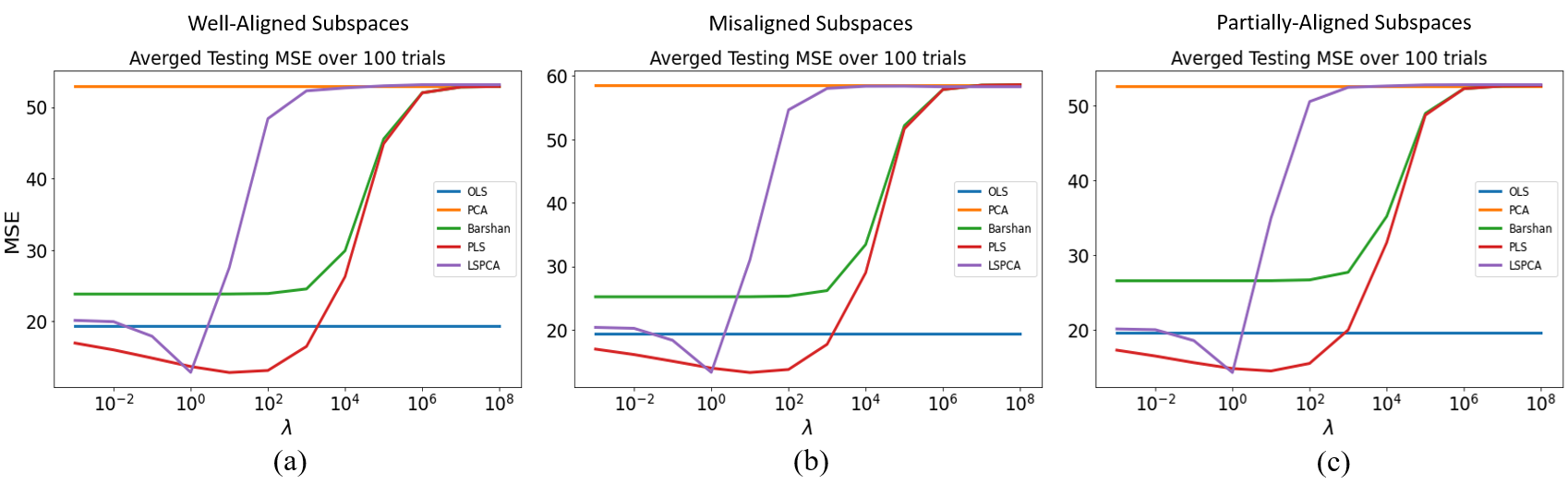}
\vspace{-5mm}
\caption{The effect of $\gamma$ on LSPCA, extended Barshan, and extended PLS. $N_{train}=150$ and the eigenvalues of $\mC$ decay slowly. (a) the case of well-aligned subspaces, (b) the case of misaligned subspaces, and (c) the case of partially-aligned subspaces.}
\vspace{-3mm}
\label{fig:gamma_effect}
\end{figure}

Three of the algorithms (LSPCA, extended Barshan, and extended PLS) use a hyperparameter $\gamma$ to balance between the supervised and the unsupervised objectives. We explore how the value of $\gamma$ affects this balance in the simulation tests. Figure \ref{fig:gamma_effect} shows the results when $N_{train}=150$ and the eigenvalues of $\mC$ decay slowly. 

 Figure \ref{fig:gamma_effect} clearly shows the impact of tuning $\gamma$ in all three subspace alignment cases for all algorithms. As $\gamma$ value increases, the performance of all three algorithms approaches the classic PCA's as expected. As $\gamma$ decreases, the supervised components play a bigger role. Different algorithms utilize the label information differently and their performance converge to different levels. The performance of LSPCA becomes very close to the performance of OLS. The small number of training samples causes OLS to overfit. Both extended Barshan and extended PLS achieve the best performance with the right balance between the unsupervised and supervised objectives. Similar balancing effect of $\gamma$ is also observed under other simulation test settings.

\section{Comparison of the methods using two real data sets} \label{sec:test_real}
We compared the performance of the reviewed algorithms using two real data sets on the UCI repository \cite{Dua:2019}: wine-quality \cite{cortez2009modeling} and Parkinsons-telemonitoring \cite{tsanas2009accurate}. The wine-quality dataset uses physicochemical features of the wine to predict its sensory quality. The Parkinsons-telemonitoring uses biomedical voice measurements to predict the clinicians' scores of Parkinson's disease symptom on the UPDRS scale. As in Section \ref{sec:test_sim}, for Barshan and PLS, the extended version is used.

\begin{figure}[htb]
\center
\includegraphics[width=\columnwidth]{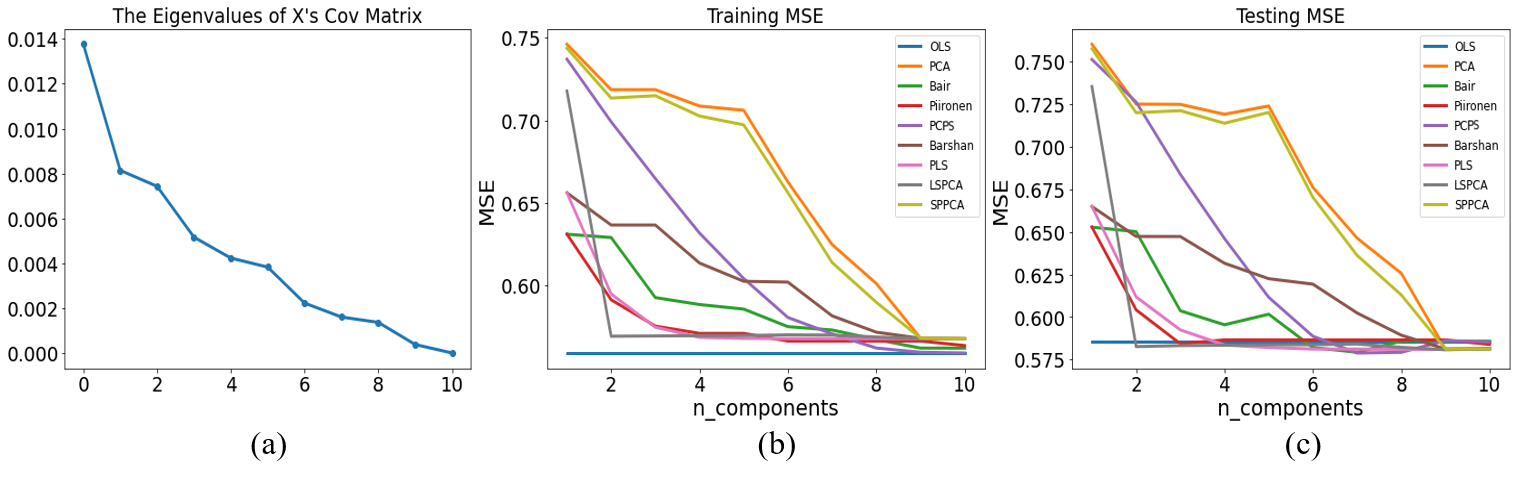}
\vspace{-5mm}
\caption{Test results on the wine-quality dataset: (a) the eigenvalues of the covariance matrix $\mC$,(b) training MSE vs. subspace dimension, and (c) testing MSE vs. subspace dimension. Different
curves correspond to different methods.}
\label{fig:wine}
\end{figure}

For the wine-quality dataset, we scaled each of the 11 features to between 0 and 1. We split the dataset to 3919 training samples and 979 testing samples. We gradually increase $K$, the dimension of the subspace to be learned, and observe the change of the model performance. The eigenvalues of the covariance matrix $\mC$, the training MSEs, and the testing MSEs are shown in  Figure \ref{fig:wine}. SPPCA and PCA again have very similar performance as we observed in the simulation tests. All other supervised methods outperform the classic PCA. LSPCA, PLS and PV method are able to find proper linear subspaces of very low dimension, demonstrating their effectiveness of incorporating the label information. 

\begin{figure}[htb]
\center
\includegraphics[width=\columnwidth]{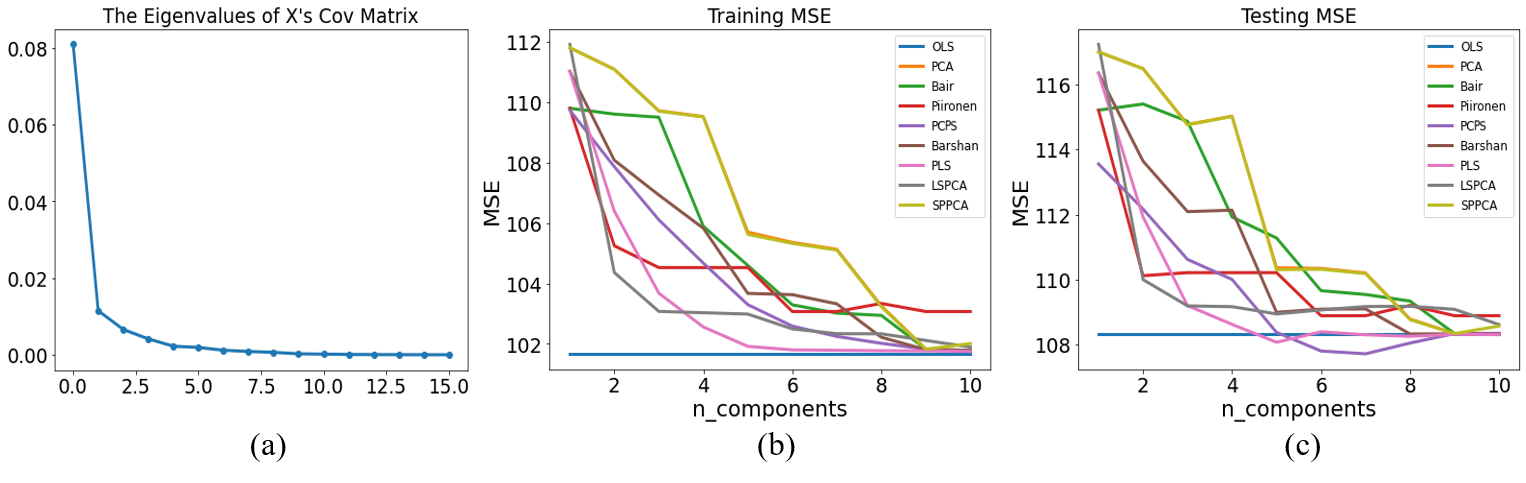}
\vspace{-5mm}
\caption{Test results on the Parkinsons-telemonitoring dataset: (a) the eigenvalues of the covariance matrix $\mC$, (b) training MSE vs. subspace dimension, and (c) testing MSE vs. subspace dimension. Different curves correspond to different methods.}
\vspace{-3mm}
\label{fig:parkinson}
\end{figure}

For the Parkinsons-telemonitoring dataset, we scaled each of the 16 medical features to between 0 and 1. We split the dataset to 4700 training samples and 1175 testing samples. The test results are shown in Figure \ref{fig:parkinson} and similar conclusions can be drawn. All supervised methods except SPPCA outperform the classic PCA. PLS, PCPS, and LSPCA are able to find proper linear subspaces of very low dimension. 

\section{Conclusion} \label{sec:conclusion}
We have reviewed several popular supervised linear dimension-reduction techniques for one-dimensional continuous response. We also extended some of these methods to include both the unsupervised and supervised components. We divided the techniques into two categories: the wrapper methods contain and the intrinsic methods. Our experimental results showed that the extended versions of PLS and LSPCA consistently outperform the other methods. Both are intrinsic methods with a balancing scheme between the supervised and unsupervised objectives.  

Since many methods contain steps to evaluate the correlation between the input variables (original or transformed) and the responses, they can also be extended to work with binary/discrete labels. Other methods such as SPPCA, PLS, and LSPCA rely on an underlying linear relationship between the transformed variables and the responses. Extending them to binary/discrete labels requires further investigation. 

Our review has focused only on linear dimension-reduction techniques. One way to extend them to non-linear models is to use the kernel trick. Moreover, all the reviewed methods can be extended to semi-supervised settings where we have both labeled and unlabled data.

\bibliography{references}

\begin{thebibliography}{14}
\providecommand{\natexlab}[1]{#1}
\providecommand{\url}[1]{\texttt{#1}}
\expandafter\ifx\csname urlstyle\endcsname\relax
  \providecommand{\doi}[1]{doi: #1}\else
  \providecommand{\doi}{doi: \begingroup \urlstyle{rm}\Url}\fi

\bibitem[Absil et~al.(2009)Absil, Mahony, and Sepulchre]{absil2009optimization}
Absil, P.-A., Mahony, R., and Sepulchre, R.
\newblock \emph{Optimization algorithms on matrix manifolds}.
\newblock Princeton University Press, 2009.

\bibitem[Bair et~al.(2006)Bair, Hastie, Paul, and
  Tibshirani]{bair2006prediction}
Bair, E., Hastie, T., Paul, D., and Tibshirani, R.
\newblock Prediction by supervised principal components.
\newblock \emph{Journal of the American Statistical Association}, 101\penalty0
  (473):\penalty0 119--137, 2006.

\bibitem[Barshan et~al.(2011)Barshan, Ghodsi, Azimifar, and
  Jahromi]{barshan2011supervised}
Barshan, E., Ghodsi, A., Azimifar, Z., and Jahromi, M.~Z.
\newblock Supervised principal component analysis: Visualization,
  classification and regression on subspaces and submanifolds.
\newblock \emph{Pattern Recognition}, 44\penalty0 (7):\penalty0 1357--1371,
  2011.

\bibitem[Cortez et~al.(2009)Cortez, Cerdeira, Almeida, Matos, and
  Reis]{cortez2009modeling}
Cortez, P., Cerdeira, A., Almeida, F., Matos, T., and Reis, J.
\newblock Modeling wine preferences by data mining from physicochemical
  properties.
\newblock \emph{Decision support systems}, 47\penalty0 (4):\penalty0 547--553,
  2009.

\bibitem[Dua \& Graff(2017)Dua and Graff]{Dua:2019}
Dua, D. and Graff, C.
\newblock {UCI} machine learning repository, 2017.
\newblock URL \url{http://archive.ics.uci.edu/ml}.

\bibitem[Geladi \& Kowalski(1986)Geladi and Kowalski]{geladi1986partial}
Geladi, P. and Kowalski, B.~R.
\newblock Partial least-squares regression: a tutorial.
\newblock \emph{Analytica chimica acta}, 185:\penalty0 1--17, 1986.

\bibitem[Hotelling(1933)]{hotelling1933analysis}
Hotelling, H.
\newblock Analysis of a complex of statistical variables into principal
  components.
\newblock \emph{Journal of educational psychology}, 24\penalty0 (6):\penalty0
  417, 1933.

\bibitem[Pearson(1901)]{pearson1901liii}
Pearson, K.
\newblock Liii. on lines and planes of closest fit to systems of points in
  space.
\newblock \emph{The London, Edinburgh, and Dublin Philosophical Magazine and
  Journal of Science}, 2\penalty0 (11):\penalty0 559--572, 1901.

\bibitem[Piironen \& Vehtari(2018)Piironen and Vehtari]{piironen2018iterative}
Piironen, J. and Vehtari, A.
\newblock Iterative supervised principal components.
\newblock In \emph{International Conference on Artificial Intelligence and
  Statistics}, pp.\  106--114, 2018.

\bibitem[Ritchie et~al.(2019)Ritchie, Scott, Balzano, Kessler, and
  Sripada]{ritchie2019supervised}
Ritchie, A., Scott, C., Balzano, L., Kessler, D., and Sripada, C.~S.
\newblock Supervised principal component analysis via manifold optimization.
\newblock In \emph{2019 IEEE Data Science Workshop (DSW)}, pp.\  6--10. IEEE,
  2019.

\bibitem[Tsanas et~al.(2009)Tsanas, Little, McSharry, and
  Ramig]{tsanas2009accurate}
Tsanas, A., Little, M., McSharry, P., and Ramig, L.
\newblock Accurate telemonitoring of parkinson’s disease progression by
  non-invasive speech tests.
\newblock \emph{Nature Precedings}, pp.\  1--1, 2009.

\bibitem[Wen \& Yin(2013)Wen and Yin]{wen2013feasible}
Wen, Z. and Yin, W.
\newblock A feasible method for optimization with orthogonality constraints.
\newblock \emph{Mathematical Programming}, 142\penalty0 (1-2):\penalty0
  397--434, 2013.

\bibitem[Wold(1975)]{wold1975soft}
Wold, H.
\newblock Soft modelling by latent variables: the non-linear iterative partial
  least squares (nipals) approach.
\newblock \emph{Journal of Applied Probability}, 12\penalty0 (S1):\penalty0
  117--142, 1975.

\bibitem[Yu et~al.(2006)Yu, Yu, Tresp, Kriegel, and Wu]{yu2006supervised}
Yu, S., Yu, K., Tresp, V., Kriegel, H.-P., and Wu, M.
\newblock Supervised probabilistic principal component analysis.
\newblock In \emph{Proceedings of the 12th ACM SIGKDD international conference
  on Knowledge discovery and data mining}, pp.\  464--473, 2006.

\end{thebibliography}
\bibliographystyle{icml2020}

\end{document}